# Solution to Quadratic Equation Using Genetic Algorithm

TANISTHA NAYAK
Department of Information Technology
National Institute of Science and Technology
Berhampur-761008, India
Email: tanisthanist213@gmail.com

TIRTHARAJ DASH
Department of Information Technology
National Institute of Science and Technology
Berhampur-761008, India
Email: tirtharajnist446@gmail.com

*Abstract*— **Solving Quadratic equation is one of the intrinsic interests as it is the simplest nonlinear equations. A novel approach for solving Quadratic Equation based on Genetic Algorithms (GAs) is presented. Genetic Algorithms (GAs) are a technique to solve problems which need optimization. Generation of trial solutions have been formed by this method. Many examples have been worked out, and in most cases we find out the exact solution. We have discussed the effect of different parameters on the performance of the developed algorithm. The results are concluded after rigorous testing on different equations.**

## I. INTRODUCTION

In the world of Computation, solving nonlinear algebra like quadratic equation and solving solution to nonlinear system of equation is the foundation of many scientific programming. Many practical problems can be transformed into the nonlinear system of equation and also we can solve the problem. However, the technique of genetic programming is an optimization process based on the evolution [1]. Selection of a large number of candidates' solutions is processed through genetic operations such as replication, crossover and mutation [2]. In this paper, we have proposed a method of solving quadratic equation based on genetic programming (GP). Genetic Programming (GP) is a popular form of evolutionary computing. Several similar approaches like Genetic Algorithms, Evolutionary strategies with genetic programming use principles and ideas of biological evolution to guide the machine to desired solution. Genetic Programming (GP) is becoming the broad area to solve very complicated problems. Koza has many examples including circuit design, program design, symbolic regression, pattern recognition, and robotic control [3]. We have some fixed problem to solve or to optimise. The solution lives in some set of potential solutions. In this Computing world the search space is too much large, and that's why we need to reduce the number of examined solutions. In real life, we find potential solution as an individual in some collection. The individual who is stronger means whose fittest function is higher, that individual decides the next generation potential function. After a number of generations, the fittest member becomes the solution for the problem. Here we find the roots of quadratic equation using genetic algorithm. This algorithm can be used to solve any kind of quadratic equation by changing the fitness function f(x) and changing the length of chromosome. Our method offers close method of solutions.

## II. RELATED WORKS

### A. Schur Decomposition Method:

In Schur Decomposition Method, the quadratic equation $Ax^2 + Bx + c = 0$ gives a complete characterization of solutions in terms of the generalized Schur decomposition and also compares various numerical solution techniques. Here exactly we give a thorough treatment of functional iteration methods based on Bernoulli's method [4]. Other methods include Newton's method with exact line searches, Symbolic solution and continued fractions. We show that functional iteration applied to the quadratic matrix equation can provide an efficient way to solve the associated quadratic Eigen value problem $(\lambda 2A + \lambda B + C) x = 0$.

### B. Multivariate linear rational Expectation Model.

In this particular model, to solve a system of linear expectional difference equation (a multivariate linear rational expectations model), we use the generalized Schur form. This method is simple to understand and use. And also it is applicable to a large class of rational expectations models [5,6].

## III. PROPOSED METHODOLOGY

A stepwise method is being followed for development of this tool. This algorithm is implemented using 'C' language. The steps followed are described in the Algorithm below.

**Proposed Algorithm:**

**Step-0:** START

**Step-1:**

    1.1 Initial population = a

    1.2 Set chromosome length = b (However, chromosomes are generated with the help of random numbers using GA)

**Step-2:** Convert the binary chromosome value to its decimal equivalent using binary to decimal conversion rule. The sign bit has to be considered.

**Step-3:** Evaluate the objective function $f(x) = x^2 + nx + m$ for each chromosome as given below.
    3.1. Convert the value of the objective function into fitness.
    3.2. If f(x)=0 for a particular chromosome, that chromosome is required accurate solution. Now display the value of chromosome and STOP. Otherwise perform next generation by continuing steps 4-5.

**Step-4:** *Selection (Tournament Selection):-* Take any two chromosomes randomly and select one with minimum value of fitness for the next generation. This process has to be repeated till we get 'a' number of chromosomes.

**Step-5:** *Crossover:*
    5.1. Take chromosome *i* and *j* randomly.
    5.2. Fix the cut-Point position and randomly decide left or right crossover and interchange the bits and resulting chromosome is used in the next generation. The crossover operation generates 'a' number of new chromosomes for the next generation.

**Step-6:** Go to Step-2.

**Step-7:** STOP

Some parameters on which the performance depends are discussed below.

*A. Data Storage*

In order to solve quadratic equation, we limited to real valued variables and real valued arithmetic. Therefore our solution will definitely give real value solution.

*B. Predators*

First step to store the equation in define grammar to describe them. A valid S-expression grammar is used. A valid S-expression grammar an identifier, a constant, or an operation, each enclosed in parenthesis. An identifier is a single letter. A constant is a positive or negative floating-point number. An Operation consists of an operand followed by two expressions. An expression is an identifier, a constant, or another operation. The operands consist of the +,−, ∗,/ etc. Binary operators, which correspond to addition, subtraction, multiplication, and division, respectively. Other operators have been added, such as square-root and the square operators. In this case, the operator used for square root is &, and the square function is represented by ^. The grammar leads to store the equation in binary form. Each leaf node represents either a constant or an identifier. However evaluations of quadratic equation follows depth first search algorithm. Let us consider following S-expression.
( ÷ (+ 0.089 0.563) X). The equation stored in tree form showed in fig1.

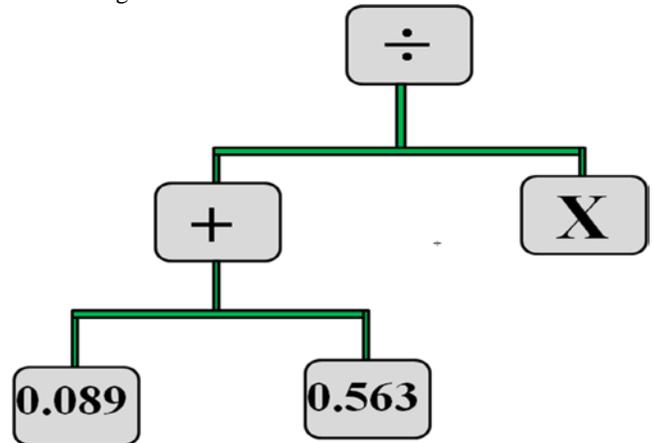

Fig1: Equation in tree format

This equation evaluates in infix notation which is given below.
(0.089 + 0.563) = 0.652. If X= 5, Then 0.652 ×5=0.652.

The standard notation is given by $\dfrac{-B \pm \sqrt{B^2 - 4AC}}{2A}$

*C. Prey*

The prey consists of the algebraic equation. For the population coefficient and possible roots are stored. Each set of each set of algebraic equations has a limited domain of coefficients in which solutions can be found. For example, linear equations are of the form Ax +B, where A, B are floating point values. This has the solution x= −B/A. This solution breaks down when A= 0. Similarly Quadratic equation has solutions in R. That type of equation can be solved by selecting randomly two real-valued roots B1 C1 coefficient multiplying out the factors. This can be given as below
(x-α) (x-β) = $x^2$ +$B_1$x +$C_1$. These equations are not static throughout the entire run of the system. During evaluation if the potential solution of the system is Є >0 then algebraic equation can be considered as *"solved"*.

The quadratic equation can be stored as following data structure given below.

typedef struct
{
    Double root1, root2; /* roots of the quadratic */
    Double A,B,C; /* coefficients of the quadratic equation
    Integer solved; /* flag show if equation was solved */
} quadraticPrey;

The flow chart of genetic algorithm can be shown in fig2.

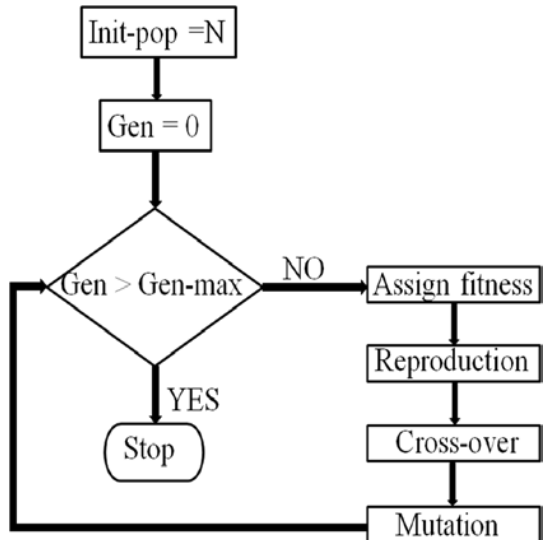

Fig2: Flowchart of Genetic Algorithm

When we use genetic algorithm, one of the important issue is determining *fitness* of the individual. An ideal fitness function guides the system in correct in correct direction. While determining the quality of a potential solution equation, simple scoring method is used. At initial state, each solution equation is allotted certain amount of hit-points. Then the equation is evaluated against a certain number of randomly chosen sets of coefficients. During the time of evaluation, a metric is taken of how well the equation did against that set of values. If the equation accurately calculated a root, few hit-points are deducted and that set of coefficients is removed from the population. If the equation did poorly, many hit-points will be subtracted away. If at any point during these competitions the equations runs out of hit-points, it is considered dead, and not allowed to reproduce. If an evaluation causes an error, by either trying to divide by zero or taking a negative square-root, the equation is also marked as dead. This method of determining fitness was used because it allowed simple, fast comparison between equations to rank them after each epoch. Also, it didn't require any knowledge of the actual solution formulas to determine a fitness for our potential solution equations. Then the success rate of evaluation can be calculated by comparing the actual root of the equation with potential root.

### D. Recombination (Crossover)

Recombination is straight forward approach to considering the structure of a solution. Before recombination begins, the surviving equations are sorted based upon their remaining hit-points. Then the equations are separated into two separate groups, highest in one group, and the lowest in another. To create a child, two random parents are chosen, with one parent from the high group and the low group being chosen. Once the two parents are selected, one is randomly chosen to be the primary parent. This parent is completely copied. Then a random node is selected as the crossover point. The other parent becomes the secondary parent. A random node is selected from this second parent, and the sub-tree from that node of the second parent is copied. Once a copy of this sub-tree has been made, it replaces the sub-tree chosen from the first tree.

### E. Mutation:

After crossover the strings are subjected to mutation. Mutation prevents the algorithms to be trapped in local minimum. Mutation plays a important role of recovering the lost genetic materials. Mutation can be considered as a background operator to maintain genetic diversity in the population. There are different kinds of mutation for different kinds of representation. Different forms are Flipping, Interchanging, Reversing, and Mutation Probability.

### IV. EXPERIMENTAL RESULT.

When the proposed algorithm was implemented to solve a particular quadratic equation $f(x) = x^2 + 2x - 7$; two important parameters affected its performance (fitness value, $f$).

1. Number of iterations
2. Number of generation

The figure given Figure 3 and 4 shows the effect of the above two parameters on the fitness of the chromosome considered.

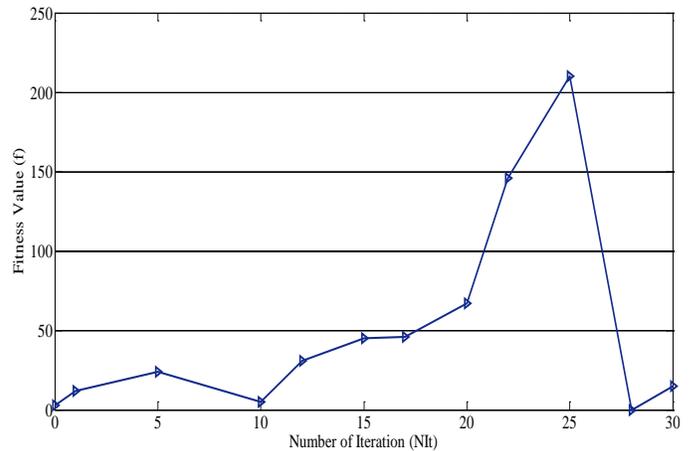

Fig3: Number of Iterations vs. Fitness of Chromosome

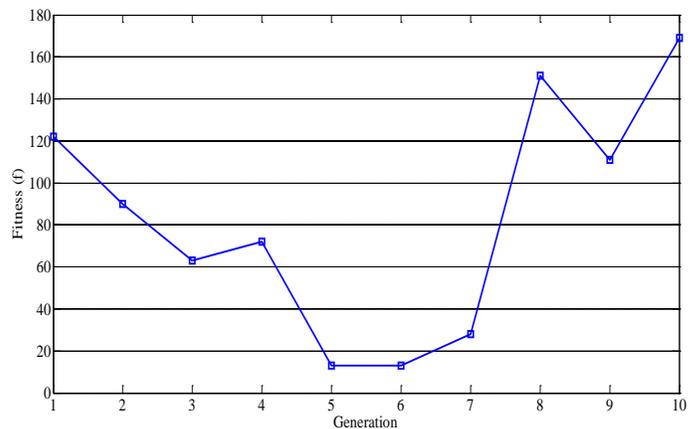

Fig4: Number Generation vs. Fitness

## V. CONCLUSION.

Genetic Programming (GP) is a very important tool for problem solving and optimization. In this paper, we proposed one algorithm to solve any quadratic equation. The number of population (a) and the chromosome length (b) can be set as per the requirement by the considered problem. The tested results showed that the fitness (f) of the chromosome depends on two important parameters (i) number of iterations and (ii) total number of generation. As future work, the approach should be made to solve a system of equation using Genetic Programming.